\documentclass{article}
\PassOptionsToPackage{numbers,sort&compress}{natbib}

 \usepackage[preprint]{neurips_2026}

\usepackage[utf8]{inputenc} 
\usepackage[T1]{fontenc} 
\usepackage{hyperref} 
\hypersetup{hidelinks}
\usepackage{url} 
\usepackage{booktabs} 
\usepackage{amsfonts} 
\usepackage{nicefrac} 
\usepackage{microtype} 
\usepackage{xcolor} 

\usepackage{xcolor} 
\usepackage{graphicx}
\usepackage{adjustbox}
\usepackage{pifont}
\usepackage{amsmath}
\usepackage{graphicx}
\usepackage{pgfplots}
\usepackage{comment}
\usepackage{amsmath,amssymb} 
\usepackage{color}
\usepackage{soul}
\usepackage{wrapfig}
\usepackage{array}
\usepackage{xspace}
\usepackage{subcaption}
\usepackage{wrapfig}
\usepackage{mathrsfs}
\usepackage{sidecap}
\usepackage{colortbl} 
\usepackage{bm}
\usepackage{enumitem}
\usepackage{xspace}
\usepackage{cutwin}
\usepackage{graphicx}
\usepackage{wrapfig} 
\usepackage{booktabs} 
\usepackage{colortbl} 

\usepackage{soul,xcolor}
\usepackage[usestackEOL]{stackengine} 

\usepackage{tcolorbox}
\usepackage{tikz}

\sethlcolor{yellow!25}

\definecolor{issuecolor}{RGB}{180,40,40} 
\definecolor{findingcolor}{RGB}{40,90,160} 

\newcommand{\method}{EverAnimate}

\newcommand{\Evae}{\mathcal{E}}
\newcommand{\Dvae}{\mathcal{D}}
\newcommand{\Epose}{\mathcal{E}_{\mathrm{pose}}}
\newcommand{\Eface}{\mathcal{E}_{\mathrm{face}}}

\title{\method{}: Minute-Scale Human Animation via Latent Flow Restoration}

\author{%
{\bfseries Wuyang Li$^{*,\dag}$ \hspace{2.2em} Yang Gao$^{*}$ \hspace{2.2em} Mariam Hassan \hspace{2.2em} Lan Feng}\\[0.1cm]
{\bfseries Wentao Pan \hspace{2.2em} Po-Chien Luan \hspace{2.2em} Alexandre Alahi$^{\dag}$}\\[0.24cm]
{\normalfont \'{E}cole Polytechnique F\'ed\'erale de Lausanne (EPFL), Switzerland}\\
{\normalfont\small Project Page:
\href{https://everanimate.github.io/homepage/}{\texttt{https://everanimate.github.io/homepage/}}}%
}

\begin{document}

\maketitle
\begingroup
\let\thefootnote\relax
\footnotetext{%
  \leftline{$^*$Equal contribution.}\\
  \leftline{\qquad $^\dag$ Correspondence to: Wuyang Li and Alexandre Alahi.}%
}
\endgroup

\begin{abstract}

We propose \textbf{EverAnimate}, an efficient post-training method for long-horizon animated video generation that preserves visual quality and character identity. Long-form animation remains challenging because highly dynamic human motion must be synthesized against relatively static environments, making chunk-based generation prone to accumulated drift: (i) low-level quality drift, such as progressive degradation of static backgrounds, and (ii) high-level semantic drift, such as inconsistent character identity and view-dependent attributes. To address this issue, \textbf{EverAnimate} restores drifted flow trajectories by anchoring generation to a persistent latent context memory, consisting of two complementary mechanisms. \emph{(i) Persistent Latent Propagation} maintains a context memory across chunks to propagate identity and motion in latent space while mitigating temporal forgetting. \emph{(ii) Restorative Flow Matching} introduces an implicit restoration objective during sampling through velocity adjustment, improving within-chunk fidelity. With only lightweight LoRA tuning, \textbf{EverAnimate} outperforms state-of-the-art long-animation methods in both short- and long-horizon settings: at 10 seconds, it improves PSNR/SSIM by 8\%/7\% and reduces LPIPS/FID by 22\%/11\%; at 90 seconds, the gains increase to 15\%/15\% and 32\%/27\%, respectively.

\end{abstract}

\section{Introduction}

Animating human characters from pose sequences is a fundamental problem in motion transfer, with broad applications in virtual avatars, content creation, and motion capture. Benefiting from the increasing capacity of video Diffusion Transformer (DiT)~\cite{wang2025wan,kong2024hunyuanvideo}, recent methods have substantially improved the realism and controllability of human animation, making it increasingly feasible to synthesize videos that are both visually plausible and temporally coherent.

Existing works~\citep{chan2018everybody,wang2018vid2vid,guo2023animatediff,xu2023magicanimate,hu2023animateanyone,kim2024tcan,zhong2024posecrafter,gan2025humandit} first extract abstracted motion representations (e.g., 2D skeletons) from videos to mitigate identity leakage and then use them to animate the reference image. Building upon this, some works focus on designing \textbf{enhanced motion representations} that incorporate cues such as depth, 3D pose, or human parsing maps~\cite{xu2023magicanimate,kim2024tcan}. Beyond motion itself, another line of research aims to enable \textbf{more flexible controls} without pose retargeting~\cite{shi2025onetoall,zhang2025steadydancer}, including animation with large body-scale differences and spatial correspondence mismatches~\cite{tan2024animate}. In addition, some works consider facial expression and audio for broader applications in short films~\cite{cheng2025wan}.

\begin{figure}[t]
\centering
\centering\includegraphics[width=1.0\linewidth]{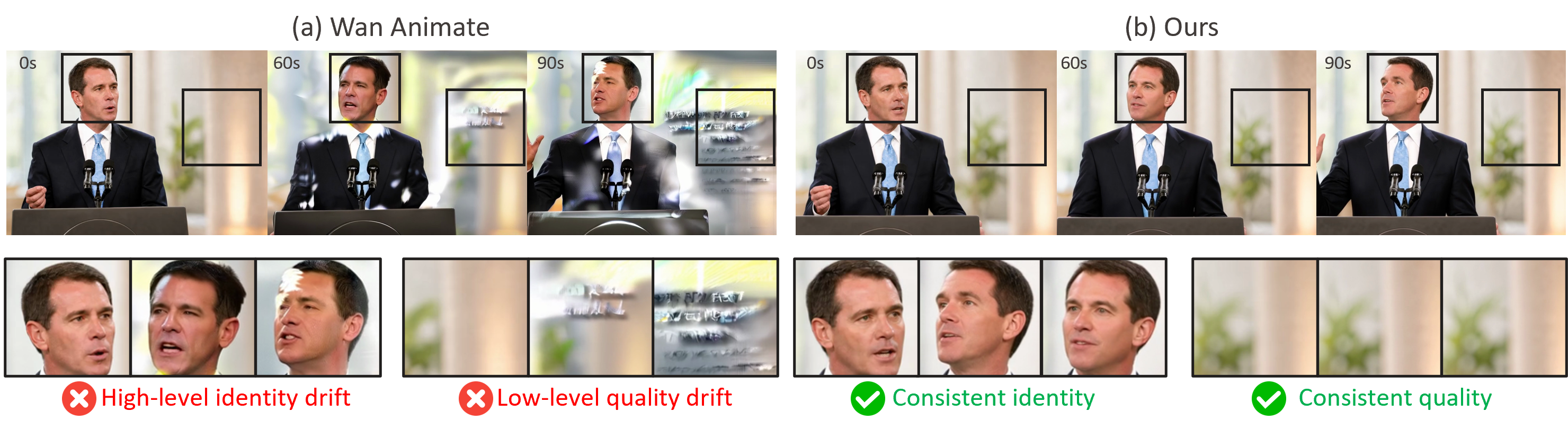}
\caption{ \textbf{(a)} Existing human animation methods primarily suffer from two types of drift: low-level quality degradation and high-level identity change. \textbf{(b)} Our method alleviates both issues, achieving stable animation. The bottom row provides zoomed-in views of the facial region and the background.}
\label{fig:intro}
\vspace{-10pt}
\end{figure}

Despite their impressive results, existing methods remain constrained to relatively short generation horizons, typically producing clips of only a few seconds. Recent works~\cite{zhang2025steadydancer, cheng2025wan} attempt to extend animation length through autoregressive, chunk-wise generation. However, the achievable extension length remains limited, only producing hundreds of frames (see Fig.~\ref{fig:qualitative}). More importantly, even with commonly adopted anti-drifting methods, such as attention sinks~\cite{xiao2023efficient}\footnote{Refer to the use of the user-provided reference frame to guide the generation of all chunks.}, error recycling~\cite{li2026stable}, and sliding-window~\cite{zhang2025steadydancer}, these approaches still accumulate errors over time and suffer from fast quality drift. Consequently, they struggle to generate minute-level animations while maintaining visual fidelity and temporal coherence throughout the entire sequence, as shown in Fig.~\ref{fig:intro}a. 

To study this issue, we begin with an intuitive observation: the core challenge of long-form animation lies in the motion heterogeneity between the background and human: \emph{Articulated human motion evolves rapidly, while much of the surrounding scene remains comparatively stable.} 
This heterogeneity causes two forms of drift (Fig.~\ref{fig:intro}a). \textbf{(i) Low-level quality drift:} Repeated cross-chunk conditioning progressively introduces and propagates texture degradation, especially in temporally stable backgrounds. \textbf{(ii) High-level identity drift:} Semantically key attributes such as identity, facial appearance, and clothing details gradually degrade, particularly under substantial motion.

To understand these issues, we conduct an empirical analysis (see Sec.~\ref{sec:motivation}), revealing two main reasons. \textbf{(i)} Repeated latent-to-pixel reconstruction progressively damages visual details during cross-chunk propagation, particularly in temporally static regions, leading to quality drift. \textbf{(ii)} Limited semantic memory, e.g., attention sinks~\cite{xiao2023efficient}, is visually helpful but insufficient for reliably anchoring long horizons, leading to identity drift. Such memory acts only as a positive signal, specifying what to preserve without identifying or correcting drift. These findings suggest a latent-space principle for stable animation: \emph{the DiT should propagate semantic memory directly in latent space across chunks, while being equipped with an intrinsic restoration ability in latent flow to correct within-chunk drift.}

Motivated by these findings, we propose \textbf{\method{}}, an efficient post-training framework for generating minute-scale long animation videos while preserving both visual quality and character identity (Fig.~\ref{fig:intro}b). \method{} introduces implicit flow restoration during latent flow propagation, which is further anchored by context memory, comprising two key components. \textbf{\emph{(i) Persistent Latent Propagation}} maintains semantic consistency across generated chunks via multi-view latent memory, thereby avoiding repeated destructive reconstruction and strengthening cross-chunk continuity. 
\textbf{\emph{(ii) Restorative Flow Matching}} enables a built-in restorative ability to actively correct emerging drift implicitly without explicitly perturbing conditional images, thereby improving within-chunk visual fidelity. With only lightweight LoRA tuning, \method{} outperforms state-of-the-art methods on both short and long generation. In summary, the contributions of this work are as follows.

\begin{itemize}
 \item We identify two major types of accumulated drift in long-form human animation, empirically reveal the limitations of image-space continuation and attention sinks (see Sec.~\ref{sec:motivation}), and derive a latent-space principle for stable long-horizon video generation.

 \item We propose \textbf{\method{}}, an efficient post-training framework for long-form human animation, built on the principle that semantic memory is propagated autoregressively across chunks, while emerging drift is corrected through intrinsic restoration.

 \item \method{} consists of two complementary components: \emph{(i) Persistent Latent Propagation}, which strengthens cross-chunk semantic continuity through latent continuation and multi-view identity memory, and \emph{(ii) Restorative Flow Matching}, which improves within-chunk visual fidelity by encouraging drift correction during sampling.

\end{itemize}

\section{Related Work}

\subsection{Human Animation} 

Early work relied on video-to-video translation~\cite{chan2018everybody,wang2018vid2vid}, with motion-transfer formulations that articulated dynamics explicitly~\cite{siarohin2019first}. Recent works solve this by first extracting intermediate motion representation, e.g., 2D poses, and then animating a reference image with image-to-video generation, which improves visual fidelity and controllability, including AnimateDiff~\cite{guo2023animatediff}, MagicAnimate~\cite{xu2023magicanimate}, and Animate Anyone~\cite{hu2023animateanyone}, as well as variants that strengthen pose conditioning and temporal consistency~\cite{zhu2024champ,zhang2024mimicmotion,tan2024animate}. Video-DiT-based approaches further scale pose-guided synthesis with unified backbones, e.g., UniAnimate-DiT~\cite{wang2025unianimatedit}, RealisDance-DiT~\cite{zhou2025realisdance}, StableAnimator~\cite{tu2025stableanimator}, and Wan-Animate~\cite{cheng2025wan}, and other works~\cite{shi2025onetoall}. Additionally, some works explore broader motion representations, e.g., 3D skeletons~\cite{yan2025scail}, multimodality~\cite{aigc_apps_VideoX_Fun_2026,gao2026deformable}, and human parsing~\cite{xu2023magicanimate,kim2024tcan}. In parallel, audio-driven body and portrait animation focuses on speech alignment and facial dynamics, with representative works like~\cite{echomimicv22024,lin2025omnihuman,chen2025hunyuanvideoavatar}. Despite strong clip-level quality, most methods focus on short horizons, and longer sequences are explored via pose-aware long generation~\cite{he2025posegen,zheng2025highfidelity}, while identity drift~\cite{seo2025lookahead} and background degradation~\cite{liu2025animateanywhere} remain challenging. In contrast, we address the train-test asymmetry in minute-scale, pose-driven animation and jointly reduce identity drift and background degradation.

\subsection{Long-form Video Generation}

Recent video foundation models have extended the effective temporal context through increasingly effective spatiotemporal compression~\cite{blattmann2023stablevideodiffusion,yang2024cogvideox,kong2024hunyuanvideo,wan2025wan,polyak2024moviegen,sand2025magi,chen2025skyreelsv2}, by scaling the model and data scale. Nevertheless, autoregressive extrapolation beyond the training horizon still suffers from a train-test mismatch, leading to exposure bias, accumulated errors, and forgetting. A complementary line of work, therefore, studies long-horizon continuation and drift control. Early methods rely on trajectory guidance or continuation heuristics~\cite{liu2024freetraj,xiao2023efficient,luansocial}. More recent approaches redesign the rollout procedure and training objective. Diffusion Forcing~\cite{chen2024diffusion} and CausVid~\cite{yin2025causvid} bridge bidirectional and autoregressive denoising. Self Forcing~\cite{huang2025selfforcing}, Rolling Forcing~\cite{liu2025rollingforcing}, and LongLive~\cite{longlive2025} mitigate exposure bias through self-conditioned rollouts and attention sink. FramePack~\cite{zhang2025framepack} breaks causality by predicting the future anchor. Recently, SVI~\cite{li2026stable}, Helios~\cite{yuan2026helios}, and Matrix-Game 3.0~\cite{wang2026matrix} enable extended generation through error restoration. LongCat-Video~\cite{meituan2025longcat} incorporates video extension during pre-training, while memory-based formulations such as MALT~\cite{yu2025malt}, PFP~\cite{zhang2025pretraining}, and WorldMem~\cite{xiao2025worldmem} preserve long-range information with latent or state-space memories. In contrast to generic long-video generation, our work targets minute-scale, pose-guided human animation, where drift arises from both the pose-conditioned motion structure and the visual synthesis process. 

\section{Preliminaries and Motivation}\label{sec:motivation}

\textbf{Problem Setup.}
Given a reference image \(I_{\mathrm{ref}}\) and a target pose-control sequence \(C_{\mathrm{pose}}=\{P_{\ell}\}_{\ell=1}^{T}\), pose-guided human animation aims to synthesize a video \(V=\{I_{\ell}\}_{\ell=1}^{T}\) that follows the target poses while preserving the identity and appearance of \(I_{\mathrm{ref}}\). Here \(P_{\ell}\) denotes the pose map at frame \(\ell\), which is distinct from the RGB frame \(I_{\ell}\). In DiT-based models, the video VAE encoder \(\Evae(\cdot)\) maps \(V\) into latent codes \(X=\Evae(V)\), and the video VAE decoder \(\Dvae(\cdot)\) reconstructs the video as \(V=\Dvae(X)\). We denote by \(G_{\theta}\) the conditional sampling procedure induced by the DiT vector field \(v_{\theta}\). Then single-clip generation can be written as \(X=G_{\theta}(\Evae(I_{\mathrm{ref}}), C_{\mathrm{pose}})\). For long-video extension, the pose-control sequence is divided into \(N\) consecutive chunks \(\{C_{\mathrm{pose}}^{(n)}\}_{n=1}^{N}\), where each chunk contains \(L\) poses. The first chunk is generated from the reference image, i.e., \(X^{(1)} = G_{\theta}(\Evae(I_{\mathrm{ref}}), C_{\mathrm{pose}}^{(1)})\). For chunk \(n \geq 2\), existing methods decode the previous latent chunk \(X^{(n-1)}\) into video, take its last carry-over frames \(I^{(n-1)}_L\), re-encode it, and use it as the carry-over condition:
\[
X^{(n)} = G_{\theta}\!\left(\Evae(I^{(n-1)}_L), C_{\mathrm{pose}}^{(n)}\right), \quad
V^{(n-1)} = \Dvae(X^{(n-1)}).
\] We present the single-frame carry-over case for simplicity, which can be extended to sliding windows.

\begin{figure}[t]
\centering
\centering\includegraphics[width=1.0\linewidth]{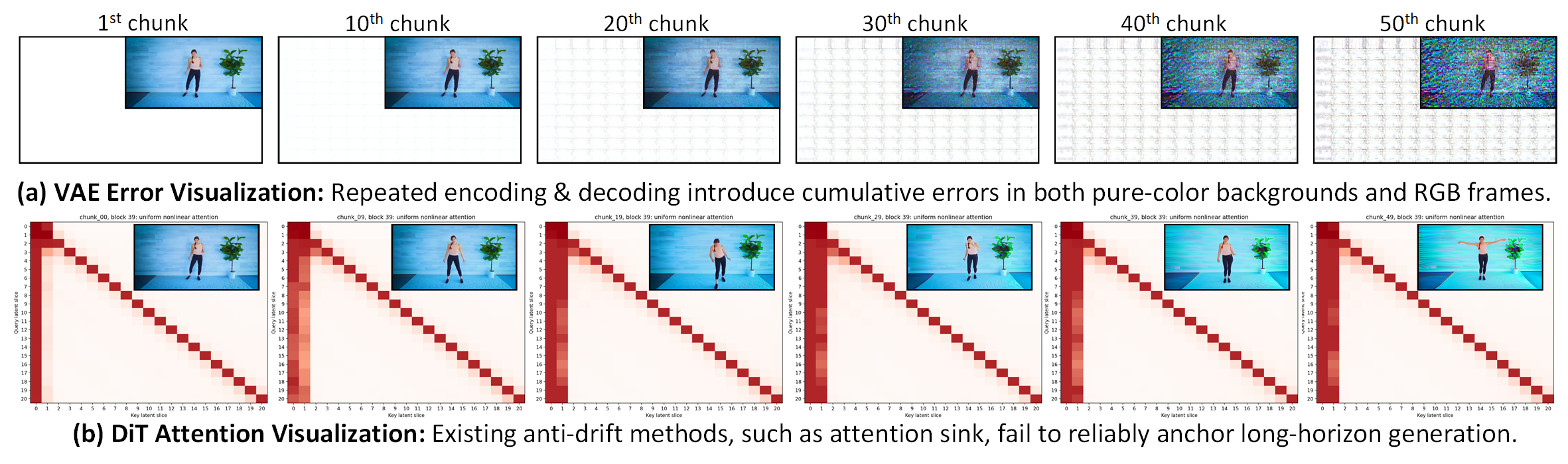}
\caption{\textbf{Illustration of errors in long-range animation videos.}
We visualize (a) VAE round-trip reconstructions and (b) DiT self-attention maps across video chunks at different lengths. In each attention map, the 1$^{st}$ \emph{col.} (highlighted in red) shows how video tokens attend to the global reference frame. \emph{Although most tokens correctly attend to this reference (i.e., forming an attention sink~\cite{xiao2023efficient,longlive2025}), the generated video still exhibits progressive degradation over time, failing to anchor the long-horizon generation.} The 2$^{nd}$ \emph{col.} shows attention to the last frame of the preceding chunk. }
\label{fig:motivation}
\vspace{-10pt}
\end{figure}

\textbf{Problem Analysis.}
We analyze the state-of-the-art method Wan-Animate \cite{cheng2025wan}\footnote{Empirically, we find that Wan-Animate performs best among prior animation baselines for long-range generation because of its attention-sink design; see the qualitative comparison.} from the perspectives of VAE and DiT representations. To mitigate long-range drift, Wan-Animate introduces a persistent identity reference \(I_{\mathrm{ref}}\) as an attention sink throughout chunk-wise generation. Specifically, the \(n\)-th chunk is generated under the additional condition of \(I_{\mathrm{ref}}\):
\[
X^{(n)} = G_{\theta}\!\left(\Evae(I^{(n-1)}_{L}), C_{\mathrm{pose}}^{(n)}, \Evae(I_{\mathrm{ref}})\right),
\]
where the re-encoded carry-over frame \(\Evae(I^{(n-1)}_{L})\) provides inter-chunk continuity, and \(\Evae(I_{\mathrm{ref}})\) is a persistent anchor (sink). However, drift remains significant, revealing two key findings (see Fig.~\ref{fig:motivation}).

\textbf{Finding 1: Repeated frame-level VAE round-trips inevitably accumulate drift.}
We first consider an idealized setting in which the DiT introduces no additional error on temporally static regions, such as the background, and predicts identical residuals across chunks. Under this assumption, any degradation can be attributed solely to the standard carry-over pipeline, which repeatedly decodes the previous latent chunk, extracts the last frame, and re-encodes it for the next chunk. In Fig.~\ref{fig:motivation}, this repeated VAE round-trip causes visible degradation even for static content, including both flat-color images and realistic animation frames. The error accumulates gradually, evolving from mild color distortion to obvious visual artifacts. This observation suggests that \emph{image-space continuation is fundamentally ill-suited for long-horizon animation, and that cross-chunk semantics should instead be propagated autoregressively in latent space}. Existing bidirectional DiTs, however, do not naturally expose a persistent latent state that can be reused across chunks without additional design.

\textbf{Finding 2: Attention sinks alone cannot fully prevent semantic and visual drift.}
We test Wan-Animate by using the persistent reference image as an attention sink~\cite{xiao2023efficient} to globally anchor identity and appearance across chunks. However, as shown in Fig.~\ref{fig:motivation}(b), noticeable drift still emerges over long-horizon generation, despite most tokens correctly attending to the reference frame and forming a clear attention sink. We attribute this limitation to three factors. \textbf{(i)} A single reference frame cannot provide sufficient information, e.g., multi-view cues, required to preserve identity under substantial changes in pose and viewpoint. \textbf{(ii)} Compared with autoregressive DiTs, bidirectional DiT generation involves longer chunks and denser token interactions, which fundamentally dilute the anchoring effect of a single sink token. \textbf{(iii)} Attention sinks act only as passive reference signals: they indicate what should be preserved but lack sufficient signals to correct the trajectory once drift occurs. 

\textbf{Remark.}
These observations point to two requirements for stable long-form animation: (i) propagate cross-chunk semantics (motion/identity/appearance) directly in latent space to prevent forgetting, and (ii) actively correct drift during sampling. Accordingly, we maintain a persistent latent state with \emph{short-term} motion memory and \emph{long-term} identity memory, and we add an ODE-based trajectory restoration to progressively pull deviated states back on track. By avoiding repeated image-space carry-over frames \cite{li2026stable}, our method improves the anti-drifting and relieves inter-chunk flicker.

\begin{figure}[t]
\centering
\centering\includegraphics[width=1.0\linewidth]{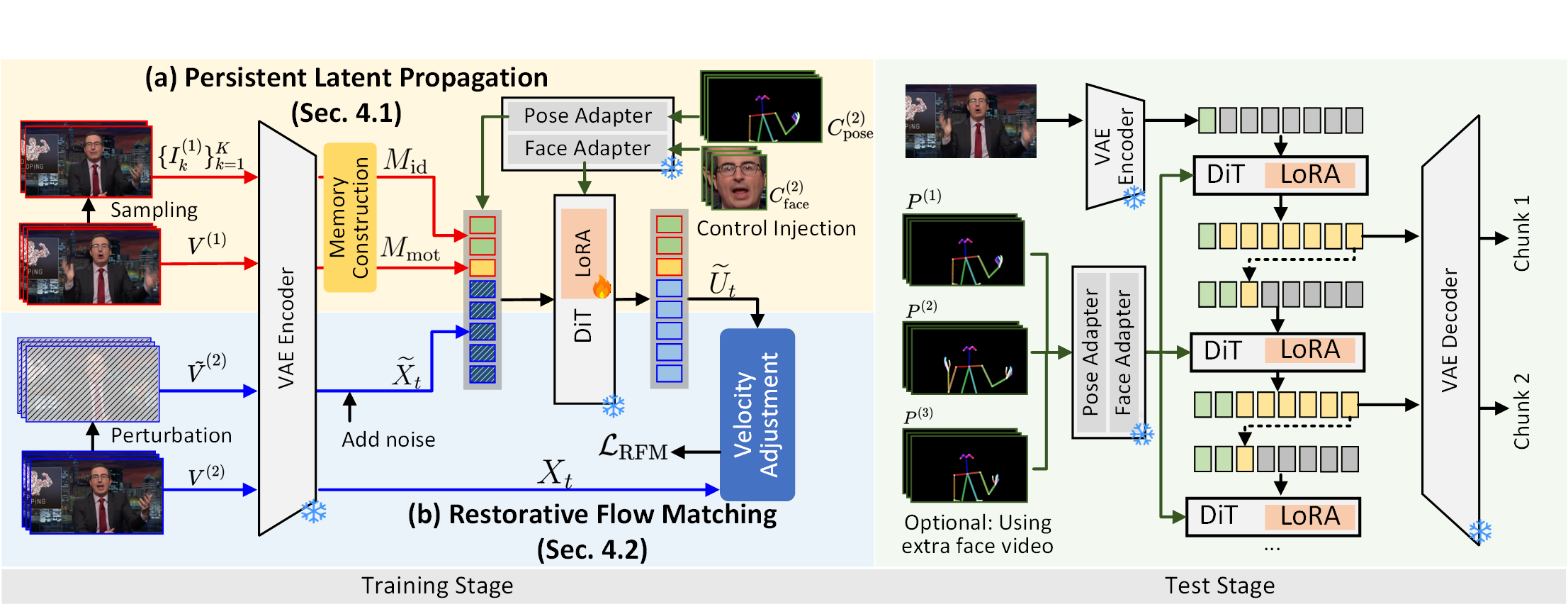}
\caption{Overview of \method{}. \textbf{Train:} from a context chunk \(V^{(1)}\), we extract motion/identity memories $M_{\mathrm{id}/\mathrm{mot}}$ and train the model to generate the next chunk \(V^{(2)}\) with restorative flow matching. \textbf{Test:} We roll out chunk-by-chunk in the latent space without decoding frames between chunks.}
\label{fig:overview}
\vspace{-10pt}
\end{figure}

\section{Method}

\label{sec:method}

\textbf{Overview.} Fig.~\ref{fig:overview} illustrates the overall workflow. During training, we use two adjacent chunks to optimize the model: \(V^{(1)}\) provides context memory $M_{\mathrm{ctx}}$ and \(V^{(2)}\) is the target chunk for the current generation. \emph{(a) Persistent Latent Propagation} constructs motion and identity memory that will be propagated from \(V^{(1)}\) to \(V^{(2)}\), preventing the high-level drift. \emph{(b) Restorative Flow Matching} encourages the generation flow to recover once the in-chunk trajectory deviates from the clean path, solving the low-level drift. In inference, after generating the current video chunk, we reuse the video latent to guide the next-chunk generation without autoregressively decoding and encoding frames.

\subsection{Persistent Latent Propagation}

Given the context chunk \(V^{(1)}\), we propose to establish a context memory $M_{\mathrm{ctx}}$ to generate \(V^{(2)}\). This consists of a motion memory $M_{\mathrm{mot}}$ that preserves short-term temporal continuity across adjacent chunks, and a global identity memory $M_{\mathrm{id}}$ that anchors multi-view identity across all chunks.

\begin{wrapfigure}{r}{0.6\textwidth}
\vspace{-10pt}
\centering
\includegraphics[width=\linewidth]{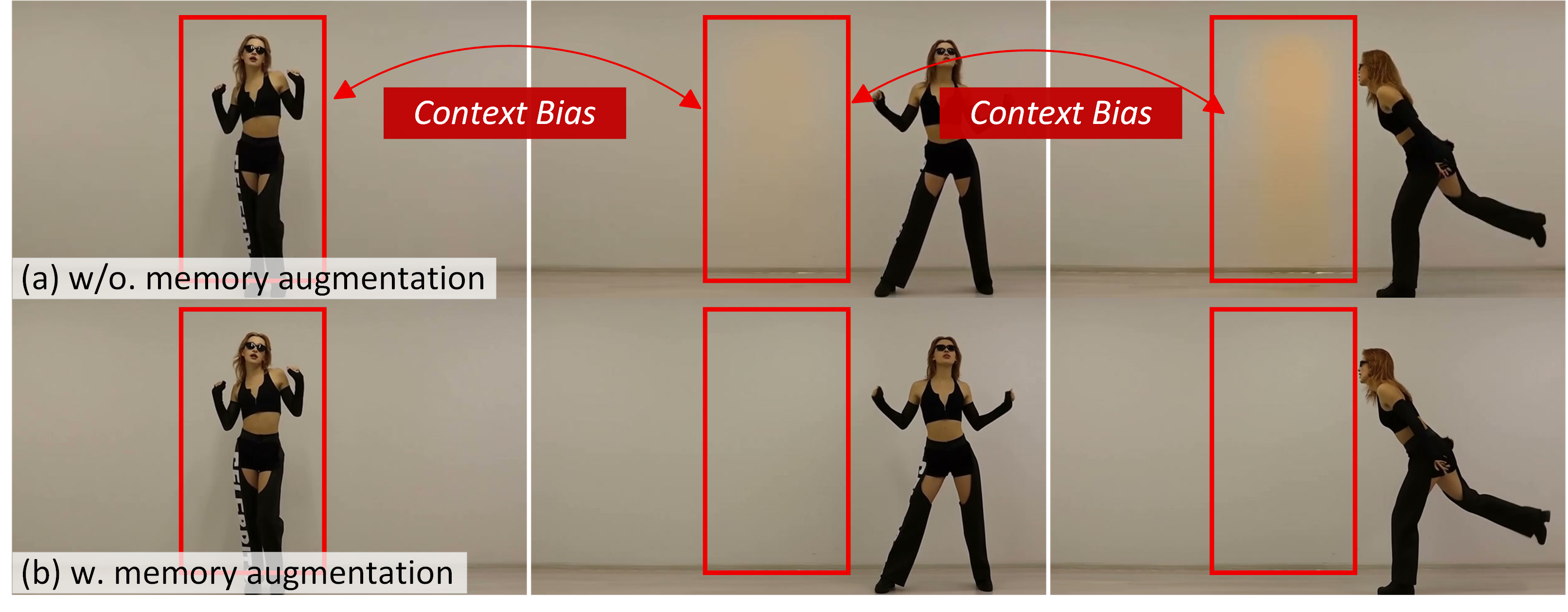}
\caption{Effects of memory augmentation \(\mathcal{T}_{\mathrm{id}}\). }
\label{fig:mem}
\end{wrapfigure}

\textbf{Memory Construction.}
Let \(X^{(1)}, X^{(2)} \in \mathbb{R}^{T_z \times H \times W \times C}\) be the clean video latents of the context chunk \(V^{(1)}\) and the target chunk \(V^{(2)}\), where \(T_z\) is the temporally compressed length produced by the video VAE $\mathcal{E}$. We extract both memories from \(V^{(1)}\). The motion memory only needs to bridge adjacent chunks, so we keep the last \(r\) latent slices. The identity memory should stay useful under pose/viewpoint changes, so we encode a small set of sampled frames:
\begin{equation}
M_{\mathrm{mot}} = X^{(1)}[T_z-r+1:T_z], \qquad
M_{\mathrm{id}} = \left\{\mathcal{E}\!\left(\mathcal{T}_{\mathrm{id}}(I_{k}^{(1)})\right)\right\}_{k=1}^{K},
\label{eq:memory-construction}
\end{equation}
where \(\{I_{k}^{(1)}\}_{k=1}^{K}=\mathrm{RandomSample}(V^{(1)}, K)\). We use random multi-view sampling so that, at test time, users can provide an arbitrary set of reference views while reducing systematic view-to-view bias in the identity memory. However, we find that directly using memory will \emph{spatially affect generation with a context bias} (see Fig.~\ref{fig:mem}a). To solve this, we propose a simple yet effective \textbf{\emph{memory augmentation}} \(\mathcal{T}_{\mathrm{id}}\) that applies mild identity-preserving spatial augmentation, e.g., random translation and rescaling, in training to prevent spatial biases of memory context. This breaks the undesirable spatial association between memory and generated frames, thereby mitigating context bias (Fig.~\ref{fig:mem}b).

\textbf{Memory and Control Injection.} We inject the memories and controls into the DiT input in two steps. First, we form the context tokens by concatenating motion/identity memories (plus a null pad to match temporal length). For Wan-style backbones, we build up the full memory as follows,
\begin{equation}
M_{\mathrm{ctx}} = \mathrm{Concat}_{t}\!\left(M_{\mathrm{mot}}, M_{\mathrm{id}}, X_{\mathrm{pad}}\right),
\label{eq:wan-style}
\end{equation}
where \(X_{\mathrm{pad}}\) is a null latent block so that \(M_{\mathrm{ctx}}\) has temporal length \(T_z\). We condition the generation of \(V^{(2)}\) on $C^{(2)} = \left\{C_{\mathrm{pose}}^{(2)}, C_{\mathrm{face}}^{(2)}\right\}$, with \(C_{\mathrm{pose}}^{(2)}=\{P_{\ell}^{(2)}\}_{\ell=1}^{L}\) aligned to \(V^{(2)}\) and \(C_{\mathrm{face}}^{(2)}\) the face guidance, following~\cite{cheng2025wan}. Second, we inject the pose into the target latent, then concatenate it with the context tokens to form the final DiT input using the pose adapter $\Epose$. This process can be written as follows,
\begin{equation}
\widehat{X}_{t}^{(2)} = X_{t}^{(2)} + \Epose\!\left(C_{\mathrm{pose}}^{(2)}\right), \qquad
H_{t}^{(2)} = \mathrm{Concat}_{\mathrm{ch}}\!\left(\widehat{X}_{t}^{(2)}, M_{\mathrm{ctx}}\right).
\label{eq:pose-injection}
\end{equation}
Face guidance is encoded by a lightweight {face adapter} \(\Eface(\cdot)\) and injected into intermediate DiT blocks via cross-attention~\cite{cheng2025wan}. For brevity, we write the conditioned vector field as \(v_{\theta}(\cdot, t \mid C^{(2)})\). In the next subsection, \(M_{\mathrm{ctx}}\) denotes the memory tokens, \(\widehat{X}_{t}^{(2)}\) is the pose-guided target latent, and \(H_{t}^{(2)}\) denotes the final DiT input. Then, the model aims to predict a velocity field that restores them residually for memory tokens, while the remaining tokens are used to generate the target video.

\begin{wrapfigure}{r}{0.5\textwidth}
\vspace{-10pt}
\centering
\includegraphics[width=\linewidth]{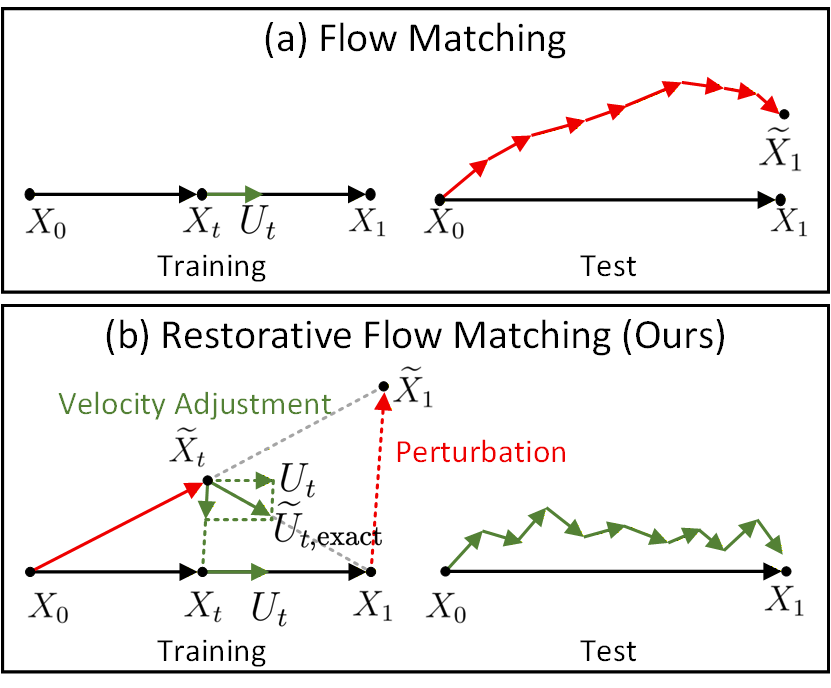}
\caption{\textbf{RFM illustration.} Comparison between our RFM and the standard FM baseline.}
\label{fig:error}
\end{wrapfigure}

\subsection{Restorative Flow Matching}
Given the memory-anchored input, we train the denoising flow not only to follow the clean flow trajectory but also to recover from small intra-trajectory deviations during rollout.

\textbf{Flow Matching (FM).}
We first recall the standard FM formulation (Fig.~\ref{fig:error}a) for the target chunk \(V^{(2)}\). Since the chunk index is fixed in this subsection, we drop the superscript \(\cdot^{(2)}\) for readability. Let \(X_0 \sim \mathcal{N}(0, I)\) denote the Gaussian source endpoint and let \(X_1=\Evae(V^{(2)})\) denote the clean latent endpoint. Standard FM defines the linear interpolant and its target velocity as follows,
\begin{equation}
X_{t} = (1-t)X_0 + t X_1, \quad
U_t = X_1 - X_0.
\label{eq:fm-clean}
\end{equation}
After memory and control injection, the DiT receives \(H_{t}^{(2)}\) from Eq.~\eqref{eq:pose-injection}. The corresponding FM objective can be written as 
\begin{equation}
\mathcal{L}_{\mathrm{FM}} =
\mathbb{E}
\left[
\left\|
v_{\theta}\!\left(H_{t}^{(2)}, t \mid C^{(2)}\right)
- U_t
\right\|_2^2
\right].
\label{eq:fm-objective}
\end{equation}
This objective trains the vector field to transport Gaussian noise toward the clean data manifold under the same memory/control pathway used at inference time. It works well when the trajectory stays close to the clean path, but autoregressive long-video generation often encounters nearby yet imperfect states that standard FM does not explicitly train the model to correct.

\textbf{From FM to Restorative FM (RFM) with Velocity Adjustment.}
During long-horizon rollout, each chunk reuses self-generated history, so small errors can propagate across chunks, leading to a drifted trajectory. Recent long-video methods~\cite{li2026stable,yuan2026helios,wang2026matrix,chen2026context} simulate this mismatch by directly perturbing the autoregressive carry-over signal. We instead keep the propagated motion latent \(M_{\mathrm{mot}}\) unchanged, simulate endpoint drift, and explicitly adjust the velocity (Fig.~\ref{fig:error}b). As a result, the model learns an intrinsic restoration ability that alleviates cross-chunk flicker. Let \(\xi\) denote a random perturbation, let \(\mathcal{T}_{\xi}\) be a perturbation operator with \(\mathcal{T}_{0}\) equal to the identity map, and define
\begin{equation}
\widetilde{V} = \mathcal{T}_{\xi}(V^{(2)}), \qquad
\widetilde{X}_{1} = \Evae(\widetilde{V}),\qquad
\widetilde{X}_{t}= (1-t)X_0 + t \widetilde{X}_{1}.
\label{eq:perturbed-endpoint-path}
\end{equation}
We then replace the target state \(X_{t}\) with \(\widetilde{X}_{t}\) while keeping the same memory and control pathways. Concretely, we form the pose-injected target latent \(\widetilde{\widehat{X}}_{t} = \widetilde{X}_{t} + \Epose(C_{\mathrm{pose}})\) and the corresponding DiT input \(\widetilde{H}_{t} = \mathrm{Concat}_{\mathrm{ch}}(\widetilde{\widehat{X}}_{t}, M_{\mathrm{ctx}})\). The model is therefore exposed only to perturbed in-chunk states, rather than perturbed transmitted context, which more effectively mitigates drift.

To derive the restorative target, we ask for the unique constant velocity that transports the current perturbed state \(\widetilde{X}_{t}\) to the clean endpoint \(X_1\) over the remaining interval \([t,1]\). Under the same linear-flow constraint used by standard FM, the continuation from time \(t\) to time \(1\) is
\begin{equation}
\widetilde{X}_{s,\mathrm{con}}=
\frac{1-s}{1-t}\widetilde{X}_{t}
+
\frac{s-t}{1-t}X_1,
\qquad s\in[t,1],
\label{eq:restoration-continuation}
\end{equation}
which satisfies \(\widetilde{X}_{t,\mathrm{con}}=\widetilde{X}_{t}\) and \(\widetilde{X}_{1,\mathrm{con}}=X_1\). Its endpoint-consistent velocity is constant:
\begin{equation}
\widetilde{U}_{t,\mathrm{exact}}
:=\frac{d \widetilde{X}_{s,\mathrm{con}}}{ds}
=
\frac{X_1-\widetilde{X}_{t}}{1-t}.
\label{eq:restoration-velocity-derivation}
\end{equation}
When \(\widetilde{X}_{t}=X_t\), namely when no perturbation is applied, this expression reduces to the standard FM velocity \(X_1-X_0\). Substituting Eq.~\eqref{eq:perturbed-endpoint-path} into Eq.~\eqref{eq:restoration-velocity-derivation} yields
\begin{equation}
\widetilde{U}_{t,\mathrm{exact}}
=
\underbrace{\bigl(X_1 - X_0\bigr)}_{U_t}
+
\underbrace{\frac{1}{1-t}\bigl(X_t-\widetilde{X}_{t}\bigr)}_{\text{restoration term}}.
\label{eq:additive-restoration}
\end{equation}
\begin{wrapfigure}{r}{0.5\textwidth}
\vspace{-10pt}
\centering
\includegraphics[width=\linewidth]{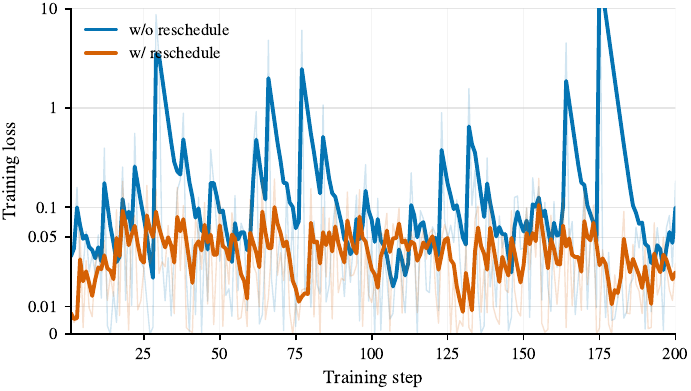}
\caption{\textbf{Effects of the reschedule.} Comparison of the training stability with and without rescheduling $\lambda{(t)}$, i.e., Eq.~\eqref{eq:additive-restoration} vs. Eq.~\eqref{eq:gaussian-restoration-weight}. Our rescheduling design, shown in orange, can stabilize training.}
\label{fig:loss}
\end{wrapfigure}
Eq.~\eqref{eq:additive-restoration} shows that RFM can be written as \emph{the standard FM velocity plus a correction that pulls the perturbed state back toward the clean path.} 

However, we find that the exact coefficient \(\frac{1}{1-t}\) is poorly conditioned and grows rapidly as \(t\rightarrow 1\) in the low-noise region, where the state is already close to the data manifold. In practice, this makes the correction term disproportionately large near the clean endpoint, leading to unstable targets, over-aggressive supervision, and potential model divergence (see the blue curve in Fig.~\ref{fig:loss}). We therefore propose to \emph{reschedule the exact coefficient} with a bounded time weight \(\lambda(t)\),
\begin{equation}
\widetilde{U}_{t}
=
U_t + \lambda(t)\bigl(X_t-\widetilde{X}_{t}\bigr),
\label{eq:restorative-velocity}
\end{equation}
where \(\lambda(t)\) follows a bounded bell-shaped schedule. A simple choice is Gaussian rescheduling 
\begin{equation}
\lambda(t)
=
\frac{
\exp\!\bigl(-\beta (t-\tfrac{1}{2})^2\bigr)
-
\exp(-\beta/4)
}{
1-\exp(-\beta/4)
},
\qquad \beta > 0,
\label{eq:gaussian-restoration-weight}
\end{equation}
which peaks in the intermediate regions and smoothly decays near both ends of the trajectory. The intuition is that both extremes are less worth correcting: \textbf{(i)} in the high-noise region, the state is dominated by Gaussian noise, so the perturbation contributes little semantic signal and heavy restoration is unnecessary; \textbf{(ii)} in the low-noise region, the noise component is small and most deviation has already been corrected by earlier steps, so additional restoration is weak and can over-constrain the clean endpoint. This makes the restorative supervision strongest where perturbations are informative and numerically well-behaved, while avoiding excessive correction near the clean endpoint to stabilize training (see Fig.~\ref{fig:loss}, orange curve). This also follows the loss-reweighting used in FM training~\cite{diffsynthstudio}. Reusing the conditioning set in Eq.~\eqref{eq:pose-injection}, we optimize
\begin{equation}
\mathcal{L}_{\mathrm{RFM}} =
\mathbb{E}\left[
\left\|
v_{\theta}\!\left(\widetilde{H}_{t}, t \mid C\right)-\widetilde{U}_{t}
\right\|_2^2
\right].
\label{eq:rfm-objective}
\end{equation}
This formulation subsumes standard flow matching: when \(\xi=0\), we have \(\widetilde{X}_{1}=X_1\), the restorative term vanishes, and \(\widetilde{U}_{t}=U_t\). When \(\xi \neq 0\), the model is trained to follow a nearby but distorted trajectory while retaining a bounded directional pull toward the clean endpoint.

\begin{figure}[t]
\centering
\includegraphics[width=1.0\linewidth]{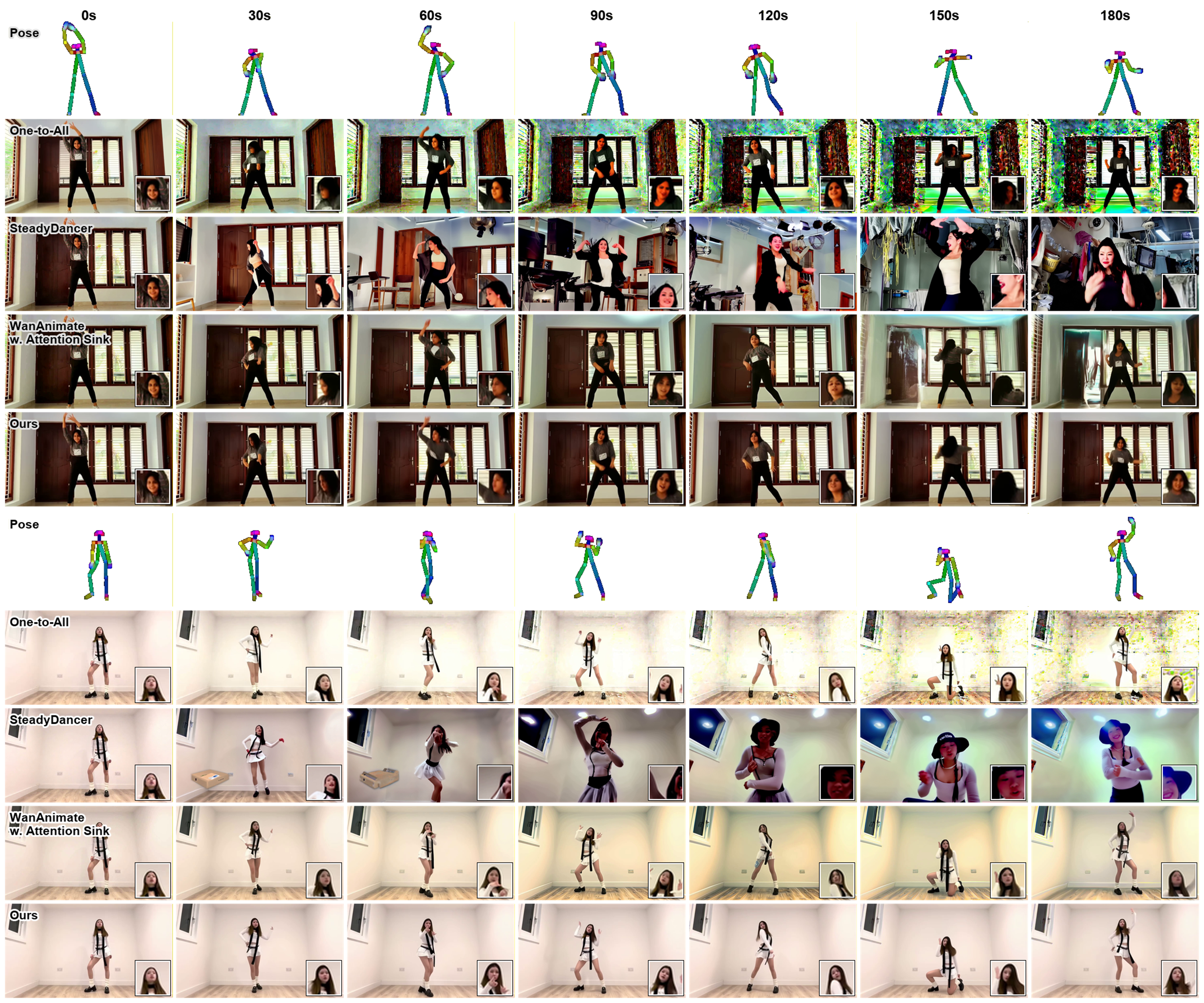}
\caption{Qualitative comparison with state-of-the-art methods. See supplementary material for videos. The bottom-right region is cropped according to the face regions in the ground-truth videos.}
\label{fig:qualitative}
\vspace{-8pt}
\end{figure}

\textbf{Remark.} Recent long-video methods (e.g., SVI~\cite{li2026stable}, Helios~\cite{yuan2026helios}, and Matrix-Game 3.0~\cite{wang2026matrix}) improve stability by perturbing the conditional image, i.e., an \emph{input-oriented} strategy. While effective for simulating error accumulation, this corrupts the carry-over condition that should remain temporally consistent across chunks, thereby undermining stability and increasing temporal flicker. We instead adopt an \emph{output-oriented} perspective: because per-step generation quality is determined by the latent state \(X_t\), we preserve the propagated motion latent and apply an explicit restorative term to the in-chunk state \(\widetilde{X}_t\). In this way, we correct the trajectory without sacrificing cross-chunk continuity.

\subsection{Training and Inference}

\noindent\textbf{Progressive Training.} Training is divided into two stages. \emph{(i) Memory adaptation.} Existing animation models are adapted from image-to-video generators, which only take a single image as the condition. We therefore first adapt the model to the memory condition by generating the current chunk using two types of memory defined in Eq.~\eqref{eq:memory-construction}. In this stage, we perturb the motion memory following~\cite{yuan2026helios,li2026stable} and optimize the model with standard FM loss, \(\mathcal{L}_{\mathrm{FM}}\).
\emph{(ii) Anti-drift adaptation.}
We then optimize the model with restorative FM loss \(\mathcal{L}_{\mathrm{RFM}}\) to improve long-range stability.
 
\noindent\textbf{Inference.} Users can flexibly provide \(m\) reference frames, where \(1 \leq m \leq K\), to specify the target identity and scene context. If \(m=K\), these reference frames are directly used as the \(K\)-frame identity memory and are shared by all chunks. If fewer reference frames are provided, e.g., in the single-reference setting, we first generate the initial chunk \(V^{(1)}\) using the available reference frame(s), and then randomly sample \(K-m\) keyframes from \(V^{(1)}\) to complete the identity memory. Once constructed, this identity memory remains fixed and is shared across subsequent chunks. For the first chunk \(V^{(1)}\), no previous motion context is available, so the motion latent \(M_{\mathrm{mot}}\) is zero-padded during generation. For each subsequent chunk \(V^{(n)}\) with \(n>1\), the last \(r\) latent slices from the previous chunk are propagated as short-term motion memory, as defined in Eq.~\eqref{eq:memory-construction}. These latent slices are directly used for conditioning, avoiding the need to decode and re-encode preceding frames.

\section{Experiments}
\label{sec:experiments}
\textbf{Implementation.} We build our method on top of Wan-2.2-Animate \cite{cheng2025wan}. During training, we set \(K=4\), corresponding to one reference frame and three additional multi-view reference images, and use \(r=1\) for short-term motion memory. The first and second training stages run for 4,000 and 1,000 iterations, respectively, with 8 GPUs. The LoRA rank and scaling factor \(\alpha\) are both set to 128. For restorative training, we randomly apply color shift, sharpness, and saturation perturbations to the target video chunk, following Helios~\cite{yuan2026helios}. During inference, we use 20 sampling steps without classifier-free guidance (CFG), using only the user-given reference frame as~\cite{cheng2025wan}.

\textbf{Datasets and Metrics.} Following One-to-All Animate~\cite{shi2025onetoall}, we train on a combination of the Champ~\cite{champ}, UBC~\cite{ubc}, and Seedance~\cite{seedance} videos and 2k self-collected minute-scale videos from Youtube with 480P resolution. For evaluation, given the lack of standardized long-video benchmarks, we report results at multiple target durations, including \(10\)s, \(30\)s, \(60\)s, and \(90\)s at 25 FPS. We benchmark generation quality from three aspects: \textbf{(i)} frame-level fidelity using PSNR/SSIM and perceptual similarity using LPIPS; \textbf{(ii)} feature-space perceptual quality at the clip level using FID and Video-MAE Distance (V-MAE, computed from VideoMAE features~\cite{tong2022videomae}); and \textbf{(iii)} long-range identity correctness and consistency using face-region PSNR (F-PSNR).

\begin{table*}[t]
\centering
\caption{Quantitative comparison with state-of-the-art methods across different temporal horizons. F-PSNR denotes the face-region PSNR used to evaluate identity correctness and consistency.}
\label{tab:benchmark}
\scriptsize
\setlength{\tabcolsep}{2.4pt}
\begin{tabular}{l|ccc|cc|c|ccc|cc|c}
\toprule
& \multicolumn{6}{c|}{\textbf{10s}} & \multicolumn{6}{c}{\textbf{30s}} \\
\cmidrule(r){2-7} \cmidrule(l){8-13}
Method & PSNR$\uparrow$ & SSIM$\uparrow$ & LPIPS$\downarrow$ & FID$\downarrow$ & V-MAE$\downarrow$ & F-PSNR$\uparrow$
& PSNR$\uparrow$ & SSIM$\uparrow$ & LPIPS$\downarrow$ & FID$\downarrow$ & V-MAE$\downarrow$ & F-PSNR$\uparrow$ \\
\midrule
One-to-All & 20.730 & 0.824 & 0.268 & 55.969 & 0.041 & 16.311 & 20.453 & 0.818 & 0.287 & 52.033 & 0.036 & 16.245 \\
SCAIL & 19.602 & 0.700 & 0.311 & 60.734 & 0.068 & 15.575 & 16.499 & 0.568 & 0.429 & 88.485 & 0.051 & 13.108 \\
SteadyDancer & 20.528 & 0.736 & 0.257 & 49.975 & 0.050 & 14.235 & 16.724 & 0.617 & 0.382 & 61.317 & 0.049 & 12.120 \\
UniAnimate-DiT & 22.773 & 0.882 & 0.192 & 45.350 & 0.044 & 16.401 & 20.908 & 0.857 & 0.238 & 58.852 & \textbf{0.032} & 15.353 \\
Wan-Animate & 23.470 & 0.839 & 0.217 & 42.898 & 0.042 & 17.278 & 21.907 & 0.787 & 0.241 & 40.809 & 0.039 & 16.426 \\
\midrule
Ours & \textbf{25.238} & \textbf{0.895} & \textbf{0.169} & \textbf{38.277} & \textbf{0.031} & \textbf{17.701}
& \textbf{24.458} & \textbf{0.873} & \textbf{0.181} & \textbf{31.190} & 0.033 & \textbf{17.650} \\
\bottomrule
\end{tabular}
\begin{tabular}{l|ccc|cc|c|ccc|cc|c}
& \multicolumn{6}{c|}{\textbf{60s}} & \multicolumn{6}{c}{\textbf{90s}} \\
\cmidrule(r){2-7} \cmidrule(l){8-13}
Method & PSNR$\uparrow$ & SSIM$\uparrow$ & LPIPS$\downarrow$ & FID$\downarrow$ & V-MAE$\downarrow$ & F-PSNR$\uparrow$
& PSNR$\uparrow$ & SSIM$\uparrow$ & LPIPS$\downarrow$ & FID$\downarrow$ & V-MAE$\downarrow$ & F-PSNR$\uparrow$ \\
\midrule
One-to-All & 20.108 & 0.802 & 0.314 & 46.951 & 0.037 & 15.495 & 19.306 & 0.754 & 0.352 & 50.354 & 0.021 & 15.492 \\
SCAIL & 12.913 & 0.419 & 0.545 & 150.123 & 0.070 & 10.846 & 11.048 & 0.349 & 0.602 & 186.992 & 0.037 & 9.937 \\
SteadyDancer & 14.344 & 0.556 & 0.478 & 81.423 & 0.064 & 11.062 & 12.997 & 0.524 & 0.541 & 101.645 & 0.043 & 10.741 \\
UniAnimate-DiT & 19.028 & 0.807 & 0.296 & 71.183 & 0.040 & 13.967 & 17.422 & 0.745 & 0.349 & 83.521 & 0.027 & 13.329 \\
Wan-Animate & 20.665 & 0.745 & 0.282 & 32.603 & 0.041 & 15.920 & 19.676 & 0.702 & 0.323 & 32.636 & 0.030 & 15.855 \\
\midrule
Ours & \textbf{23.857} & \textbf{0.855} & \textbf{0.194} & \textbf{26.241} & \textbf{0.027} & \textbf{16.927}
& \textbf{22.646} & \textbf{0.810} & \textbf{0.220} & \textbf{23.981} & \textbf{0.027} & \textbf{16.623} \\
\bottomrule
\end{tabular}
\vspace{-10pt}
\end{table*}

\subsection{Main Results}

\textbf{Qualitative Comparison.} Fig.~\ref{fig:qualitative} compares generation quality across different rollout horizons. We can find that most models produce visually plausible results at short horizons but gradually deteriorate over time. In contrast, our method can maintain stable quality in the background and human identity without obvious artifacts, demonstrating robustness for long-range generation.

\textbf{Quantitative Comparison.} Tab.~\ref{tab:benchmark} shows that our method consistently achieves the best performance across rollout horizons, compared with state-of-the-art pose animation methods~\cite{shi2025onetoall,yan2025scail,zhang2025steadydancer,wang2025unianimatedit,cheng2025wan}. Note that generated videos may exhibit camera motion that differs from the ground truth, making pixel-level metrics such as PSNR and SSIM less reliable. At 10s, it improves over Wan-Animate, increasing PSNR from 23.47 to 25.24 and reducing LPIPS from 0.217 to 0.169. The advantage becomes more pronounced at longer horizons, such as 90s, indicating the stability of our method.

\begin{wraptable}{r}{0.46\textwidth}
\vspace{-1.0em}
\centering
\caption{Ablation study at 60s (480p).}
\label{tab:ablation}
\small
\setlength{\tabcolsep}{4pt}
\begin{tabular}{lccc}
\toprule
Method & PSNR$\uparrow$ & SSIM$\uparrow$ & LPIPS$\downarrow$ \\
\midrule
Baseline & 18.472 & 0.543 & 0.386 \\
w/o RFM& 21.842 & 0.781 & 0.273 \\
w/o PLP & 22.317 & 0.809 & 0.208\\
Full model & \textbf{23.857} & \textbf{0.855} & \textbf{0.194} \\
\bottomrule
\end{tabular}
\vspace{-0.8em}
\end{wraptable}

\textbf{Ablation Study.} Tab.~\ref{tab:ablation} reports a 60s ablation. The full model improves over the baseline by +5.39 PSNR, and increases SSIM from 0.543 to 0.855. Removing RFM yields noticeably lower perceptual quality. Removing PLP primarily weakens cross-chunk carry-over, suggesting that memory propagation is important for long-horizon consistency. 

\section{Conclusion}
We propose \method{}, a lightweight post-training framework for long-form pose-guided human animation. Rather than relying on image-space continuation, \method{} addresses long-horizon degradation through latent-state control. \emph{(i) Persistent Latent Propagation} maintains reusable latent memory across chunks, preserving identity and motion cues over extended rollouts. \emph{(ii) Restorative Flow Matching} complements this mechanism with a bounded corrective step during sampling, steering perturbed latent trajectories back toward the clean generation path. Together, these designs enable more stable long-form generation and substantially improve consistency and fidelity.

\newpage
\section*{Acknowledgment}

This work was supported as part of the Swiss AI Initiative by a grant from the Swiss National Supercomputing Centre (CSCS) under project ID a144 on Alps, Sportradar, Valeo, and Honda R\&D Co., Ltd. We would like to express our gratitude to Valentin Gerard, Adrien Lefevre, Zimin Xia, and the Longcat Team for insightful discussions.

\bibliographystyle{unsrtnat}
\bibliography{references}

\newpage
\appendix

\section{Discussion}
\subsection{Limitations}
\label{sec:limitations}

Our method substantially improves temporal robustness for minute-scale pose-guided human video generation under a fixed, chunked rollout setting. However, it does not completely eliminate long-horizon drift in some cases: \textbf{errors may still appear in certain samples} when generation is extended to extremely long lengths, especially under out-of-distribution appearances. This might be caused by the limited capacity of LoRA training and data diversity. A further limitation is that the current framework \textbf{does not provide explicit control over camera viewpoint or camera trajectory}. The relative motion between the subject and the camera is modeled only implicitly and is largely determined by the training distribution and stochastic generation process. As a result, the method may have limited controllability when users require precise camera movements, consistent viewpoint changes, or coordinated subject-camera motion over long durations.

Future work may further improve scalability, reduce ultra-long-horizon error accumulation, and introduce explicit camera-control mechanisms for specifying viewpoint, camera trajectory, and subject-camera relative motion. It would also be valuable to evaluate the method across a broader range of identities, clothing styles, camera motions, scene layouts, and real-world usage scenarios.

\subsection{Broader Impact}
\label{sec:broader-impact}

Long-form human animation can support a variety of beneficial applications, including creative content production, virtual avatars, education, accessibility, and assistive communication. By improving the temporal consistency of synthesized human motion, the proposed approach may lower the cost and technical barriers of producing character animation and personalized visual media.

\subsection{Safeguards and Responsible Release}
\label{sec:safeguards}

Given the dual-use potential of high-fidelity human animation generation, we do not support unrestricted deployment. Any release of models, code, or demonstrations should follow a responsible process that may include gated access, explicit use policies, audit logging, and prohibitions on impersonation, non-consensual synthetic media, harassment, and politically deceptive content. Where practical, technical measures should also be adopted, including visible or invisible watermarking, and compatibility with content authentication standards. In addition, each release should provide clear documentation of intended applications and relevant ethical considerations.

\subsection{Statistical Significance}
Due to the high computational cost of long-horizon video generation and evaluation, we do not include repeated-run confidence intervals or statistical significance tests for all settings. We acknowledge this as a limitation. Nevertheless, we report results consistently across methods under the same evaluation setup, and the observed improvements are supported by both quantitative metrics and qualitative long-form comparisons.

\subsection{Ethics and Societal Impacts}
Sections~\ref{sec:broader-impact} and~\ref{sec:safeguards} discuss the dual-use nature of the proposed technology, including both beneficial applications and potential misuse risks. We explicitly recommend consent, provenance, watermarking, access control, and policies against impersonation and non-consensual synthetic media for any real-world deployment or release. When we open-source the project, we will further strengthen these safeguards through responsible release practices and technical protections wherever feasible.

\end{document}